\documentclass{article}

\usepackage[preprint]{neurips_2023}

\usepackage{svg}

\usepackage{graphicx} 
\usepackage[utf8]{inputenc} 
\usepackage[T1]{fontenc}    
\usepackage[hidelinks]{hyperref}       
\usepackage{booktabs}       
\usepackage{amsfonts}       
\usepackage{nicefrac}       
\usepackage{microtype}      
\usepackage{xcolor}         
\usepackage{amsmath}
\usepackage{listings}
\usepackage{algorithm2e}
\usepackage{caption}
\usepackage{subcaption}
\usepackage{tikz}
\usepackage{cleveref}
\usepackage{listings}

\usepackage{multirow}
\title{Contextualized Topic Coherence Metrics\thanks{\url{https://github.com/hamedR96/CTC}}}

\author{
  Hamed Rahimi$^*$\thanks{\href{mailto:hamed.rahimi@sorbonne-universite.fr}{hamed.rahimi@sorbonne-universite.fr}} \\
  Sorbonne University\\
  Paris, France \\
     \And
   Jacob Louis Hoover \\
   McGill University \\
   Montreal, Canada \\
    \And
   David Mimno \\
  Cornell University\\
  Ithaca, NY, USA \\
     \AND
   Hubert Naacke \\
  Sorbonne University\\
  Paris, France 
     \And
   Camelia Constantin \\
  Sorbonne University\\
  Paris, France
       \And
     Bernd Amann$^*$ \\
  Sorbonne University\\
  Paris, France \\
}

\begin{document}
\maketitle
\def\thefootnote{*}\footnotetext{These authors contributed equally to this work}

\begin{abstract}
The recent explosion in work on neural topic modeling has been criticized for optimizing automated topic evaluation metrics at the expense of actual meaningful topic identification.
But human annotation remains expensive and time-consuming.
We propose LLM-based methods inspired by standard human topic evaluations, in a family of metrics called Contextualized Topic Coherence (CTC).
We evaluate both a fully automated version as well as a semi-automated CTC that allows human-centered evaluation of coherence while maintaining the efficiency of automated methods.
We evaluate CTC relative to five other metrics on six topic models and find that it outperforms automated topic coherence methods, works well on short documents, and is not susceptible to meaningless but high-scoring topics.

\end{abstract}

\section{Introduction}
Topic models are a family of text-mining algorithms that identify themes in a large corpus of text data \citep{blei2012probabilistic,churchill2022evolution}. These models are widely used for exploratory data analysis with the aim of organizing, understanding, and summarizing large amounts of text data \citep{abdelrazek2022topic}.
Numerous techniques, algorithms, and tools have been employed to develop a variety of topic models for different tasks and purposes \citep{srivastava2017autoencoding,thompson2020topic,zhao2021topic} including much recent work on neural topic models \citep{grootendorst2022bertopic,angelov2020top2vec,rahimi2023antm}.
However, due to their nature as unsupervised models, comparing topic outputs, hyperparameter settings, and overall model quality has traditionally been difficult \citep{harrando2021apples,hoyle2022neural,doogan-buntine-2021-topic}.

Topic Coherence (TC) metrics measure the interpretability of topics generated by topic models. These metrics are categorized into two classes: automated TC metrics and human-annotated TC metrics \citep{hoyle2021automated}. Automated TC metrics estimate the interpretability of topic models with respect to various factors such as co-occurrence or semantic similarity of topic words. On the other hand, human-annotated TC metrics are protocols for designing surveys that rate or score the interpretability of topic models. Human judgment is often used to validate topic coherence metrics to provide an accurate assessment of the semantic coherence and meaningfulness of a given set of topics. \citep{NewmanDavid2009Eeot,aletras2013evaluating,mimno2011optimizing}. While human-annotated TC metrics incorporate subjective human judgments and provide a more accurate and nuanced understanding of how well topic models are performing (e.g. in terms of their ability to capture the underlying themes in a text corpus), they are expensive, time-consuming, and require multiple human-subjects to avoid personal biases. On the other hand, automated metrics are more cost-effective than human-annotated methods, as they do not require the hiring and training of human annotators, which results in their ability to evaluate large amounts of data and iterate through many model comparisons.

Topic Models were initially evaluated with held-out perplexity as an automated metric \citep{blei2003latent}. Perplexity quantifies how well a statistical model predicts a sample of unseen data and is computed by taking the inverse probability of the test set, normalized by the number of words in the dataset. According to \citep{chang2009reading}, perplexity has been found to be inconsistent with human interpretability. As a result, the field shifted towards adopting automated topics coherence metrics that rely on word co-occurrence-based methods like Point-wise Mutual Information (PMI) \citep{cover1999elements}. These metrics were introduced because they were believed to align more closely with human judgment, providing a better measure of the interpretability of topic words. Recently, however, several studies have called this correlation into question. For example, \citep{doogan-buntine-2021-topic} studied automated evaluation metrics and examined the validity of coherence measures. They claimed the target of interpretability is ambiguous and concluded that current automated topic coherence metrics are unreliable for evaluating topic models in short-text data collections and may be incompatible with newer neural topic models. In a similar study, \citep{hoyle2021automated} also has shown that topics generated by neural models are often qualitatively distinct from traditional topic models while they receive higher scores from current automated topic coherence metrics. \citep{hoyle2021automated} have concluded that the validity of the results produced by fully automated evaluations, as currently practiced, is questionable, and they only help when human evaluations cannot continue the comparison. These studies bring the validity of automated metrics into attention and question whether automated metrics are consistent with human judgments of topic quality. \citep{hoyle2022neural} in another recent work has shown that neural topic models fail to improve on the traditional topic models such as Gibbs LDA \citep{griffiths2004finding} in MALLET \citep{mccallum2002mallet} and considered neural topic broken as they do not function well for their intended use.

Hence, there is a demand for new automated coherency measures that are context-aware and can handle neural topic models and short-text datasets. To address these problems, we introduce   Contextualized Topic Coherence (CTC) metrics which are a context-aware family of topic coherence metrics based on the pre-trained Large Language Models (LLM). Taking Advantage of LLMs elevates the understanding of language at a very sophisticated level incorporating its linguistic nuances, contexts, and relationships. CTC is much less susceptible to being fooled by meaningless topics that often receive high scores with traditional topic coherence metrics. This paper presents two approaches using LLMs for defining CTC metrics: The first approach uses LLMs to compute contextualized estimates of the pointwise mutual information (CPMI) between topic words. In the second approach, we use existing chatbots such as ChatGPT \citep{openai2022chatgpt} to evaluate topic coherence similar to human-annotated metrics.

\paragraph{Contributions.} First, we introduce the notion of CTC metrics as a context-aware family of topic coherence metrics based on the pre-trained Large Language Models (LLM). Second, we will deliver a comprehensive analysis of the results showing the validity of CTC compared to traditional topic coherence metrics in a series of experiments with a group of six topic models that include recent advanced neural topic models (ETM \citep{dieng2020topic}, ATM \citep{wang2019atm}, CTM \citep{bianchi-etal-2021-pre}, BERTopic \citep{grootendorst2022bertopic}, Top2Vec \citep{angelov2020top2vec}) and a dominant traditional topic model (Gibbs LDA \citep{griffiths2004finding}) on two different datasets. Third, we will re-evaluate these models using the proposed metrics and show that CTC metrics work well on short documents, and are not susceptible to meaningless but high-scoring topics.

\section{Automated Topic Coherence Metrics}
Topic coherence (TC) metrics measure the consistency of words in a given topic to evaluate the interpretability and meaningfulness of a topic by computing the level of semantic similarity among words that are included in the topic. A high TC value indicates that the words in the topic are semantically similar and are likely to co-occur in the same circumstances. 
 
The authors of \citep{NewmanDavid2009Eeot,10.1145/1816123.1816156} claim that a method based on the Point-wise Mutual Information (PMI) gives the largest correlations with human ratings. They define UCI, which measures the strength of the association between pairs of words based on their co-occurrence in a sliding window of length-$l$ words. Topic coherence over PMI ($\text{TC}_{\text{UCI}}$) is defined as the average of the $\log_2$ ratio of co-occurrence frequency of word $w_i^r$ and $w_i^s$ within a given topic $i$.
\begin{equation}
\label{tcuci}
\text{TC}_{\text{UCI}} = \frac{1}{n}\sum_{i=1}^{n}\frac{1}{\binom{m}{2}} \sum_{r=2}^{m}\sum_{s=1}^{r-1}\text{PMI}(w_{i}^r, w_{i}^{s})
\end{equation}
 with
 \begin{equation}
 \label{eq:pmi}
\text{PMI}(w^i, w^j) = \log_2 \frac{P(w^i, w^j)+\epsilon}{P(w^i) P(w^j)}
\end{equation}
where $n$ is the number of topics with $m$ topic words and $\text{PMI}$ represents the pointwise mutual information between each pair of words ($w_i^r$ and $w_i^s$) in the topic $i$. $\text{PMI}$ is computed by taking the logarithm of the ratio of the joint probability of two words $P(w_i^r, w_i^s)$ appearing together to the individual probabilities of the words $P(w_i^r)$, $P(w_i^s)$ occurring separately. Note that $\epsilon=1$ is added to avoid the logarithm of zero. 
 
\citep{mimno2011optimizing} proposes UMass, an asymmetric confirmation measure that estimates the degree of coherence between words within a given topic by calculating the $\log$ ratio frequency of their co-occurrences in the corpus of documents. UMass counts the number of times a pair of words co-occur in a given corpus and compares this number to the expected number of co-occurrences where words are randomly distributed across the whole corpus. More formally, UMass computes the co-document frequency of word $w_i^r$ and $w_i^s$ divided by the document frequency of word $w_i^s$.

\begin{equation}
\label{tcumass}
\text{UMass}(w_i^r, w^s_i) = \log \frac{D(w_i^r, w^s_i) + \epsilon}{ D(w_i^s)}
\end{equation}

where $n$ and $m$ are the numbers of topics and topic words respectively. The smoothing parameter $\epsilon$ was initially introduced to be equal to one and avoid the logarithm of zero.

\citep{aletras2013evaluating} proposes context vectors for each topic word $w$ to generate the frequency of word co-occurrences within windows of $\pm$1 words surrounding all instances of $w$. They showed that NPMI \citep{bouma2009normalized} has a larger correlation with human topic ratings compared to UCI and UMass. Additionally, NPMI takes into account the fact that some words are more common than others and adjusts the frequency of individual words accordingly\citep{lau2014machine}.
 \begin{equation}
 \text{NPMI}(w_i^r, w^s_i) = \frac{\log_2 \frac{P(w_i^r, w^s_i)+\epsilon}{P(w_i^r) P(w^s_i)}}{- \log_2 (P(w_i^r, w^s_i)+\epsilon)}
\end{equation}
While NPMI is generally more sensitive to rare words and can handle small datasets, UMass focuses on fast computation of coherence scores over large corpora. \citep{stevens2012exploring} showed that a smaller value of $\epsilon$ tends to yield better results than the default value of $\epsilon=1$ used in the original paper since it emphasizes more the word combinations that are completely unattested. 

\citep{roder2015exploring} proposes a unifying framework of coherence measures that can be freely combined to form a configuration space of coherence definitions, allowing their main elementary components to be combined in the context of coherence quantification. For example, they propose the $\text{C}_\text{V}$ metric, which uses a variation of NPMI to compute topic coherence over a sliding window of size $N$ and adds a weight $\gamma$ to assign more strength to more related words. According to \citep{campagnolo2022topic}, the $\text{C}_\text{V}$ metric is more sensitive to noisy information and dirty data than $\text{C}_{\text{UMass}}$ and $\text{C}_{\text{UCI}}$. 
 \begin{equation}
  \text{C}_{\text{V}}(w_{i}^r, w_{i}^{s}) = \text{NPMI}^\gamma(w_{i}^r, w_{i}^{s}) 
 \end{equation}

\citep{10.1145/2911451.2914720} and \citep{schnabel2015evaluation} propose the metric $\text{TC}_{\text{DWR}}$ based on the Distributed Word Representations (DWR)~\citep{mikolov2013distributed,mikolov2013efficient} which are better correlated to human judgment. One way to estimate $\text{TC}_{\text{DWR}}$ is to compute the average pairwise cosine similarity between word vectors in a topic as follows.
 \begin{equation}
 \label{eq:tcdwr}
\text{DWR} (w_{i}^r, w_{i}^{s})= \frac{{w^{r}_i \cdot w^{s}_i}}{{\|w^{r}_i\| \cdot \|w^{s}_i\|}}
  \end{equation}

Similarly, \citep{ramrakhiyani2017measuring} presents a coherence measure based on grouping topic words into buckets and using Singular Value Decomposition (SVD) and integer linear programming-based optimization to create coherent word buckets from the generated embedding vectors. \citep{korenvcic2018document} proposes several topic coherence metrics based on topic documents rather than topic words. The approach essentially extracts topic documents, vectorizes them using several methods such as word embedding aggregation, and computes a coherence score based on the document vectors. \citep{lund2019automatic} proposes an automated evaluation metric for local-level topic models by introducing a task designed to elicit human judgment and reflect token-level topic quality.

\section{Contextualised Topic Coherence}
\label{ttq}

In this article, we introduce Contextualized Topic Coherence (CTC) to refer to a new family of topic coherence metrics that benefit from the recent development of Large Language Models (LLM). 

\subsection{Automated CTC}
\paragraph{CPMI.} Recent work by \citep{hoover2021linguistic} uses conditional PMI estimates to analyze the relationship between linguistic dependencies and statistical dependencies between words. They propose Contextualized PMI (CPMI) as a new method for estimating the conditional PMI between words \textit{in context} using a pre-trained language model. The CPMI between two words $w_i$ and $w_j$ in a sentence $s$ is defined as
\begin{equation}
\label{eq:cpmi}
\text{CPMI}(w_i,w_j\mid s)=\log\frac{p(w_i \mid s_{-w_i})}{p(w_i \mid s_{-w_{ij}})}
\end{equation}
where $p$ is an estimate for the probability of words in context based on a pre-trained masked language model (MLM), such as BERT. Here, $s_{-w_i}$ represents the sentence with word $w_i$ masked, and $s_{-w_{ij}}$ is the sentence with both words $w_i$ and $w_j$ masked. 

\begin{figure}[htbp]
  \centering
  \includegraphics[width=0.8\linewidth]{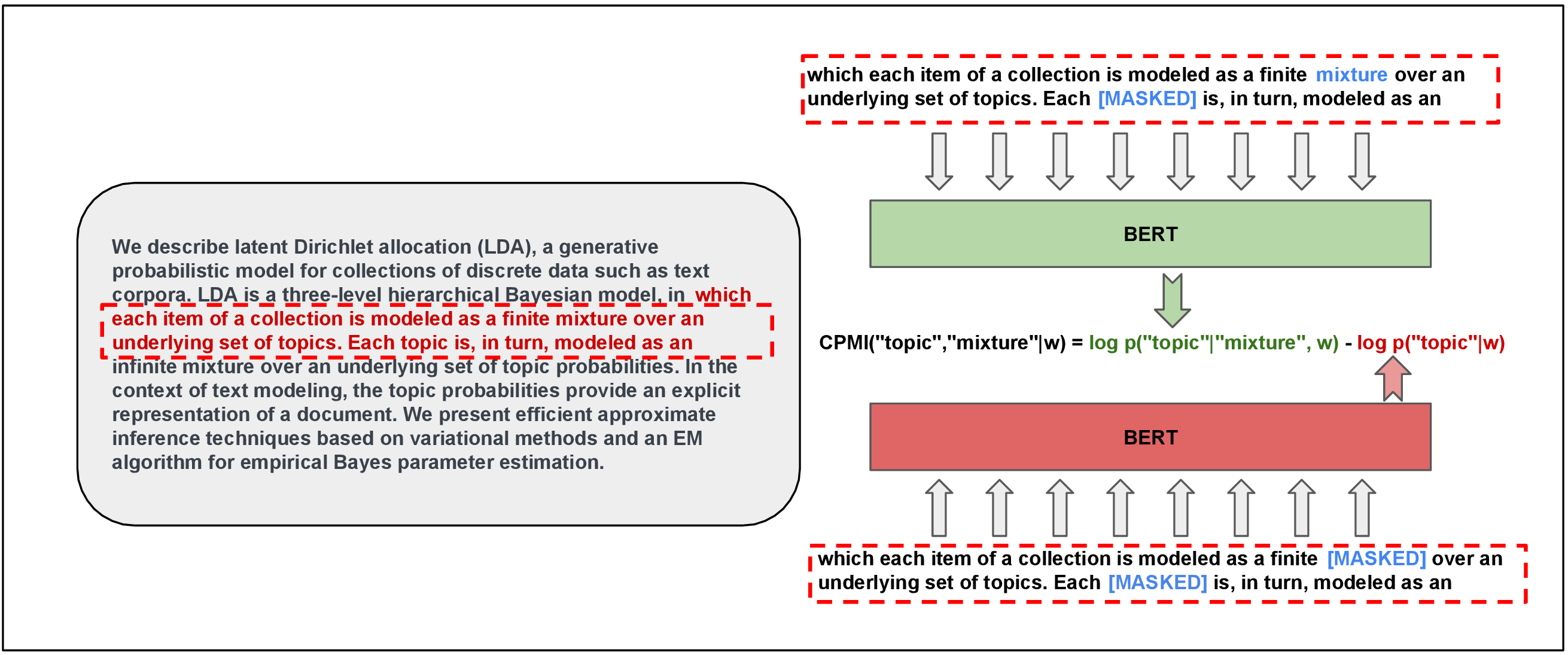}
  \caption{Calculating CPMI for two topic words in a segment of a document. The red box will slide over the whole documents and calculate CPMI for each pair of topic words.}
  \label{fig:cpmi}
\end{figure}
We adopt CPMI for introducing a new automated Contextualized Topic Coherence (CTC) metric. \Cref{fig:cpmi} illustrates the computation of automated CTC, which estimates statistical dependence within a topic in a corpus by calculating CPMI between every pair of topic words within a sliding window. 
Therefore, the first step in this procedure is to split the corpus into a set of window segments with a length of $w$ that have $k$ words intersection with adjacent window segments. Afterward, we compute the CPMI between each pair of words within each topic, and average over all the window segments, giving the following expression for CTC:
\begin{equation}
 \text{CTC}_{\text{CPMI}} = \frac{1}{n}\sum_{i=1}^{n}\frac{1}{\binom{m}{2}} \sum_{r=2}^{m}\sum_{s=1}^{r-1}\text{CPMI}(w_{i}^r, w_{i}^{s}\mid c^u)
 \label{equ:cpmi}
\end{equation}
where $c^u\subset \text{corpus}\;D$ is a window segment with length of $w$ that has $k$ words overlapping with its adjacent window segments, $n$ is the number of topics and $m$ is the number of topic words.

\subsection{Semi-automated CTC}
\paragraph{Intrusion.} \citep{chang2009reading} studied the \textit{topic words intrusion} task to assess topic coherence by identifying a coherent latent category for each topic and discovering the words that do not belong to that category. These \textit{intruder words} are detected by human subjects to assess the quality of topic models and to measure a coherence score that takes into account a low probability for intruder words to belong to a topic. We adopt this notion to chatbots with the following prompt, which provides the topic words to ChatGPT\citep{openai2022chatgpt} and asks for a category and intruder words.
\begin{center}
\small
\begin{verbatim}
      I have a topic that is described by the following keywords:[topic_words].
       Provide a one-word topic based on this list of words and identify all 
    intruder words in the list with respect to the topic you provided. Results be 
      in the following format: topic: <one-word>, intruders: <words in a list>
\end{verbatim}
\end{center}
The number of intrusion words ($\left|I_i\right|$) returned by chatbot for each topic $i$, is used to define 
CTC$_{\text{Intrusion}}$ as follows:
\begin{equation}
\label{eq:intrusion}
   \text{CTC}_{\text{Intrusion}}= \sum_{i=1}^{n} \frac{1-\frac{\left|I_i\right|}{m} }{n}
\end{equation}
where $n$ is the number of topics and $m$ is the number of topic words.

\paragraph{Rating.} While human topic ratings are expensive to produce, they serve as the gold standard for coherence evaluation \citep{roder2015exploring}. For example, \citep{syed2017full} uses human ratings to explore the coherence of topics generated by LDA topics across full texts and abstracts. \citep{newman2010automatic} provide human annotators with a rubric and guidelines for judging whether a topic is useful or useless. The annotators evaluate a randomly selected subset of topics for their usefulness in retrieving documents on a particular topic and score each topic on a 3-point scale, where 3=highly coherent and 1=useless (less coherent). Following \citep{newman2010automatic}, \citep{aletras2013evaluating} presented topics without intruder words to Amazon Mechanical Turk to score them on a 3-point ordinal scale. We adapt this idea to chatbots with the following prompt, which provides the topic words to ChatGPT and asks to rate the usefulness of the topic words for retrieving documents on a given topic. The CTC$_{\text{Rating}}$ for a topic model is then obtained by the average sum of all ratings over all the topics. 
\begin{center}
\small
\begin{verbatim}
      I have a topic that is described by the following keywords: [topic_words]. 
      Evaluate the interpretability of the topic words on a 3-point scale where
       3=“meaningful and highly coherent”  and 0=“useless” as topic words are 
      usable to search and retrieve documents about a single particular subject. 
      Results be in the following format: score: <score> 
\end{verbatim}
\end{center}

\section{Experiments}
\label{experiment}
In this section, we anticipate observing the limitations of traditional metrics in assessing neural topic models, as well as their inability to effectively handle short-text datasets, as demonstrated in~\citep{doogan-buntine-2021-topic,hoyle2021automated}. This implies that baseline metrics often yield high scores for incoherent topics, while conversely assigning low scores to well-interpretable topics. In contrast, CTC has a better model of language and can better evaluate topical similarity \textit{as it would appear to a human reader}. Therefore, we expect to see that baseline metrics and CTC would differ at extremes of highest or lowest coherency. In next section, we will investigate these assumptions using a meta-analysis of topic words and scores from different topic models and coherence metrics.

\subsection{Experimental setup}
\paragraph{Datasets.} Two datasets have been chosen for the experiment. The first dataset is 20Newsgroups dataset by \citep{Lang95} that is a collection of approximately 20K newsgroup documents, partitioned evenly across 20 different newsgroups describing various topics. This dataset has been extensively studied with many papers published on topic modeling since it provides a diverse set of documents across multiple topics, making it an ideal test bed for evaluating topic modeling algorithms. The second dataset is a collection of 17K tweets by Elon Musk published between 2017 and 2022 by \citep{raza2023elon}. Each tweet includes a text message, the date and time it was posted, and other metadata such as the number of likes and retweets. The purpose of selecting the second dataset is to evaluate the performance of topic models on short-text data, specifically in terms of their ability to produce coherent topics as measured by baseline and CTC metrics.

\paragraph{Topic Models.} The experiments involve six different topic models including traditional topic models that are widely used and the new neural topic models that have been recently proposed. The first topic model is Gibbs LDA \citep{griffiths2004finding} which is a variation of Latent Dirichlet Allocation (LDA)~\citep{blei2003latent}. 
\citep{hoyle2021automated} show that Gibbs LDA is still hard to beat with respect to human evaluation. The second topic model is a generative topic model called Embedded Topic Model (ETM) \citep{dieng2020topic}. The generative process in ETM is similar to LDA but enhanced with word embeddings, where a topic is drawn according to a probability distribution for each document. The third topic model is a neural topic model based on Generative Adversarial Networks (GANs) that is called Adversarial-neural Topic Models (ATM)~\citep{wang2019atm}. ATM  generates word-level semantic representations with Dirichlet prior and uses a generator network to capture the semantic patterns among latent topics. The fourth topic model is an algorithmic neural topic model called Top2Vec \citep{angelov2020top2vec}, which 
generates joint word-document embedding with Doc2Vec~\citep{le2014distributed}, clusters document embedding vectors in lower dimension space, and finally, represents each cluster with words that have a low cosine distance to the centroid of each cluster in the original space. 
The fifth topic model is a neural topic model based on ProdLDA \citep{srivastava2017autoencoding} and calle Contextualized Topic Model (CTM)~\citep{bianchi-etal-2021-pre}. CTM extends ProdLDA's auto-encoder with contextualized representations of documents from pre-trained large language models. The final topic model is BERTopic \citep{grootendorst2022bertopic}, an algorithmic neural topic model that generates document embeddings with pre-trained language models (BERT) and extracts topic representation throughs a class-based variation of TF-IDF. In the case of parametric models, we selected the number of topics based on the values reported by their original papers. Specifically, for the 20Newsgroup dataset, we chose 20, 50, and 100 topics, while for Elon Musk's Tweets, we selected 10, 20, and 30 topics. We maintained the remaining settings at default values as suggested by the main papers.

\paragraph{Topic Coherence Metrics.} The topics generated by the topic models are evaluated using the proposed Contextualized Topic Coherence (CTC) metrics, which are then compared to the well-established automated topic coherence metrics C$_{\text{V}}$, UCI, UMass, NPMI, and DWR. 
By comparing the results of our proposed evaluation metric with those of the established baseline metrics, we can gain valuable insights into the performance of automated topic evaluation metrics. 
For $\text{CTC}_\text{CPMI}$, we segmented the 20Newsgroup and Elon Musk's Tweets datasets into chunks of 15 and 20 words, respectively, without intersections. We then extracted the CPMI for all word pairs in each segment using the pre-trained language models \textit{bert-base-uncased} and \textit{Tesla K80 15 GB GPU} from Google Colab \citep{bisong2019google}. This pre-computing step took about 7 hours, however, afterward, one can computing can compute CTC$_\text{CPMI}$ for any topic model in the order of a few seconds. The advantage of pre-computing CPMI between all word pairs is that one can run it on very large datasets and open-source it for CTC$_\text{CPMI}$ calculation. Note that calculating CPMI only for the word pairs of all topics takes on the order of a few minutes. For evaluating $\text{CTC}_\text{Intrusion}$ and $\text{CTC}_\text{Rating}$, we made a request for each topic to \textit{ChatGPT} with \textit{GPT 3.5 Turbo}, which cost less than a dollar for all the experiments.

\subsection{Results}

\begin{table}[htb]
\scriptsize
\centering
\caption{The Results of Topic Coherence Metrics on 20Newsgroup dataset.} 
\label{tab:20Newsgroup}
\begin{tabular}{ |l|l|lllll||lll| }
\hline
\multicolumn{2}{ |c| }{Topic Models} & \multicolumn{5}{ |c|| }{Baseline Metrics} & \multicolumn{3}{ |c| }{CTC Metrics}\\
\hline
 & \#T & UCI & UMass & NPMI & C$_V$ & DWR & Rating & Intrusion & CPMI \\ 
\hline
\hline
\multirow{3}{*}{Gibbs LDA \citeyearpar{blei2003latent}} 
& 20  & \textit{0.260} & \textit{-2.338} & \textit{0.043}  & \textit{0.512} & \textit{0.211}  & \textit{1.3}  & 0.225          & \textit{0.058}$\times e^{-6}$\\
& 50  & -0.121         & -2.771          & 0.023           & 0.479          & 0.191           & 1.16          & 0.220          & 0.035$\times e^{-6}$\\
& 100 & -0.690         & -3.030          & 0.002           & 0.450          & 0.149           & 1.14          & \textit{0.267} & 0.019$\times e^{-6}$\\ 
\hline
\multirow{3}{*}{ETM \citeyearpar{dieng2020topic}} 
& 20  & \textit{0.478} & -2.08                    & \textit{0.067} & \textit{0.563} & 0.292          & 0.7                     & \textit{\textbf{0.452}} & 0.112$\times e^{-6}$\\
& 50  & 0.380          & \textit{\textbf{-1.903}} & 0.054          & 0.532          & \textit{0.330} & 1.22                    & 0.348                   & 0.119$\times e^{-6}$\\
& 100 & 0.351          & -1.962                   & 0.049          & 0.522          & 0.312          &  \textit{1.23}          & 0.41            & \textit{\textbf{0.132}}$\times e^{-6}$\\ 
\hline
\multirow{3}{*}{ATM \citeyearpar{wang2019atm}} 
& 20  & -1.431          & -3.014          & -0.059          & 0.338            & \textit{0.151} &  0.92          &  0.305         & 0.001$\times e^{-8}$\\
& 50  & -0.940          & -2.902          & -0.046          & 0.342            & 0.077          &  \textit{1.15} & 0.275          & 0.001$\times e^{-6}$\\
& 100 & \textit{-0.735} & \textit{-2.741} & \textit{-0.032} & \textit{0.362}   & 0.053          &  1.12          & \textit{0.340} & \textit{0.010}$\times e^{-6}$\\ 
\hline
\multirow{3}{*}{CTM \citeyearpar{bianchi-etal-2021-pre}} 
& 20  & -1.707          & -4.082          & 0.005             & \textit{0.601} & \textit{0.268} & 1.25            & 0.385          & \textit{0.042}$\times e^{-6}$\\
& 50  & \textit{-0.724} & \textit{-3.008} & \textit{0.046}    & 0.590          & 0.236          & \textit{1.56}   & 0.380          & 0.041$\times e^{-6}$\\
& 100 & -0.926          & -3.118          & 0.027             & 0.561          & 0.210          & 1.31            & \textit{0.392} & 0.036$\times e^{-6}$\\ 
\hline
Top2Vec \citeyearpar{angelov2020top2vec} & 85 & \textit{\textbf{0.910}} & \textit{-2.449} & \textit{\textbf{0.192}} & \textit{\textbf{0.785}} & \textit{\textbf{0.473}} &  \textit{\textbf{1.670}} & \textit{0.399} & \textit{0.022}$\times e^{-6}$\\ 
\hline
BERTopic \citeyearpar{grootendorst2022bertopic} & 145 & \textit{-1.023} & \textit{-5.033} & \textit{0.098} & \textit{0.681} & \textit{0.309} &  \textit{1.517} & \textit{0.359} & \textit{0.017}$\times e^{-6}$\\
\hline
\end{tabular}
\end{table}
\begin{table}[htb]
\scriptsize
\centering
\caption{Topic Coherence of recent NTMs on Elon Musk's Tweets dataset}
\label{tab:EM_Tweets}
\begin{tabular}{|l|l|lllll||lll|}
\hline
\multicolumn{2}{ |c| }{Topic Models} & \multicolumn{5}{ |c|| }{Baseline Metrics} & \multicolumn{3}{ |c| }{CTC Metrics}\\
\hline
& \#T & UCI & UMass & NPMI & C$_\text{V}$ & DWR & Rating & Intrusion & CPMI \\ 
\hline
\hline
\multirow{3}{*}{Gibbs LDA \citeyearpar{blei2003latent}} 
& 10 &\textit{-0.441} &\textit{-3.790} & \textit{0.016} &\textit{0.498} &\textit{0.838} &\textit{1.6}   & 0.29   &\textit{0.059}$\times e^{-6}$ \\
& 20 & -1.834 & -5.415 & -0.049 & 0.395 & 0.798 & 1.5   &  0.225 & 0.028$\times e^{-6}$\\
& 30 & -3.068 & -6.390 & -0.099 & 0.336 & 0.783 & 1.466 &\textit{0.33}   & 0.023$\times e^{-6}$\\ 
\hline
\multirow{3}{*}{ETM \citeyearpar{dieng2020topic}} 
& 10 &\textit{\textbf{0.205}} & -3.209 &\textit{\textbf{0.051}} &\textit{0.560} & 0.952 & 1.1   &\textit{0.24}  &\textit{\textbf{0.146}}$\times e^{-6}$ \\
& 20 & 0.155 &\textit{\textbf{-3.079}} & 0.028 & 0.538 & 0.974 &\textit{1.433} & 0.233 & 0.121$\times e^{-6}$\\
& 30 & 0.025 & -3.215 & 0.022 & 0.515 &\textit{\textbf{0.978}} & 1.05  & 0.195 & 0.116$\times e^{-6}$\\ 
\hline
\multirow{3}{*}{ATM} 
& 10 & -9.021 & -12.859 & -0.324 &\textit{0.364} & 0.730 &\textit{1.2}   & 0.211 & -0.001$\times e^{-6}$\\
& 20 & -7.967 & -11.770 & -0.283 & 0.343 & 0.694 & 1.1   & 0.177 &\textit{0.0001}$\times e^{-6}$\\
& 30 &\textit{-7.278} &\textit{-11.301} &\textit{-0.258} & 0.350 &\textit{0.753} & 0.933 &\textit{0.214} & -0.006$\times e^{-6}$\\ 
\hline
\multirow{3}{*}{CTM \citeyearpar{bianchi-etal-2021-pre}} 
& 10 &\textit{-2.614} &\textit{-7.049} &\textit{-0.030} &\textit{\textbf{0.580}} &\textit{0.888}  &\textit{\textbf{2.0}}   &\textit{\textbf{0.439}} & 0.027$\times e^{-6}$\\
& 20 & -3.720 & -8.336 & -0.070 & 0.534 & 0.880  & 1.45  & 0.185 & \textit{0.082}$\times e^{-6}$\\
& 30 & -3.589 & -8.063 & -0.064 & 0.573 & 0.873  & 1.766 & 0.276 & 0.069$\times e^{-6}$\\ 
\hline
Top2Vec \citeyearpar{angelov2020top2vec} & 164 & \textit{-6.272} & \textit{-10.536} & \textit{-0.152} & \textit{0.401} & \textit{0.847} & \textit{1.481} & \textit{0.274} & \textit{0.016}$\times e^{-6}$\\ 
\hline
BERTopic \citeyearpar{grootendorst2022bertopic} & 217 & \textit{-4.131} & \textit{-11.883} & \textit{-0.020} & \textit{0.432} & \textit{0.541} & \textit{1.539} & \textit{0.276} & \textit{0.011}$\times e^{-6}$\\
\hline
\end{tabular}
\end{table}
\Cref{tab:20Newsgroup,tab:EM_Tweets} represent the results of the evaluation of the topic models obtained from the 20Newsgroup and Elon Musk's Tweets datasets, respectively, using CTC and the baseline metrics. To allow us to compare the models in terms of topic coherence metrics, the highest value for each metric is shown in bold. On the other hand, the highest values for each metric within each topic model are noted in \textit{italic} font, which helps us to identify the optimal parameter (e.g., number of topics) for each model except Top2Vec and BERTopic, which are non-parametric. 

\begin{figure}[htb]
  \centering
  \begin{subfigure}[b]{0.45\textwidth}
  \centering
    \includegraphics[width=\textwidth]{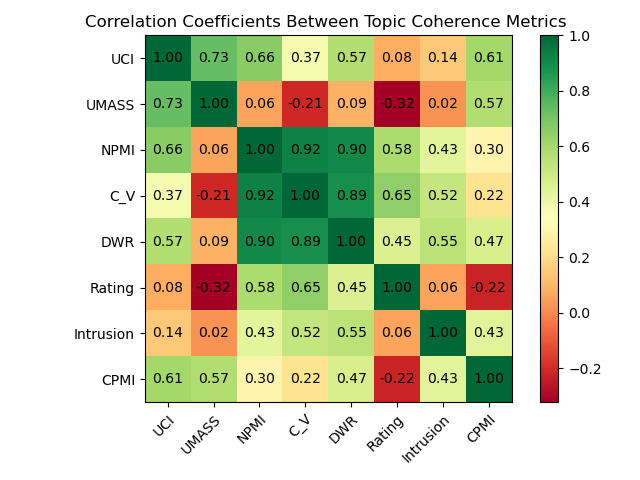}
    \caption{20Newsgroup}
    \label{fig:cor-a}
  \end{subfigure}
   \hfill
  \begin{subfigure}[b]{0.45\textwidth}
  \centering
    \includegraphics[width=\textwidth]{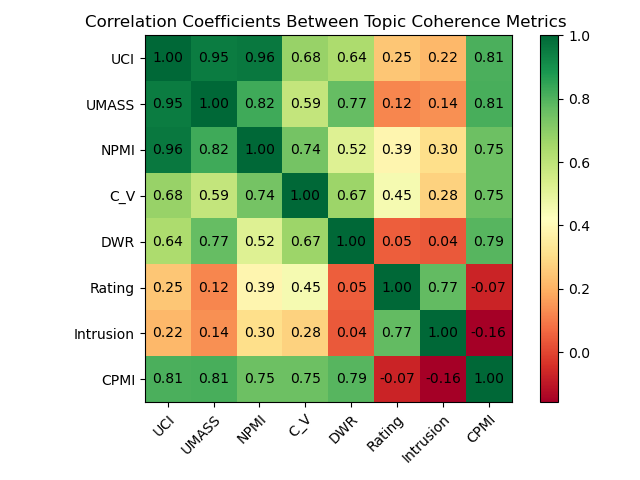}
    \caption{Elon Musk Tweets}
    \label{fig:cor-b}
  \end{subfigure}
  \caption{Pearson's correlation coefficient on CTC and baseline}
\label{fig:cor}
\end{figure}
Before analyzing the numbers, we continue to examine the relationship between CTC metrics and baseline metrics by performing Pearson's correlation coefficient analysis \citep{sedgwick2012pearson} on the results from \Cref{tab:20Newsgroup,tab:EM_Tweets} similar to \citep{doogan-buntine-2021-topic}. As shown in \Cref{fig:cor-a}, for 20Newsgroup, the baseline measures UCI and UMass are highly correlated with CPMI but not with $\text{CTC}_\text{Rating}$ and $\text{CTC}_\text{Intrusion}$, which are more correlated with the baseline measures NPMI and C$_V$ and DWR (which are also highly correlated). 
On the other hand, for the short-text EM Tweets dataset, \Cref{fig:cor-b} shows that CPMI has a high correlation with all baseline methods, whereas  $\text{CTC}_\text{Intrusion}$ and $\text{CTC}_\text{Rating}$ 
are completely independent of CPMI and the baseline measures. 

\paragraph{Observations.} As expected, \Cref{tab:20Newsgroup} reports that the baseline metrics (except for UMass) point to Top2Vec while CTC metrics (except for $\text{CTC}_\text{Rating}$) point to ETM for achieving the highest scores. Similarly, \Cref{tab:EM_Tweets} reports that the baseline metrics (except for C$_{\text{V}}$) point to ETM while CTC metrics (except for $\text{CTC}_\text{CPMI}$) point to CTM for achieving the highest scores. 
These contradictions between CTC and baseline metrics are aligned with our expectations and we will explore them with a meta-analysis of topics generated by these topic models and the scores they have received from CTC and baseline metrics. 

\paragraph{Meta-analysis.}


To estimate the consistency of the scores obtained by different coherence metrics, we will compare the coherence of high and low-scoring topics from different topic models and CTC metrics. Note that CTC metrics observe contextual patterns between topic words, and therefore, we expect them to provide more accurate coherence scores according to the interpretability of the generated topics for all topic models. 

\begin{figure}[htbp]
\centering
\begin{subfigure}[b]{0.45\textwidth}
\includegraphics[width=\textwidth]{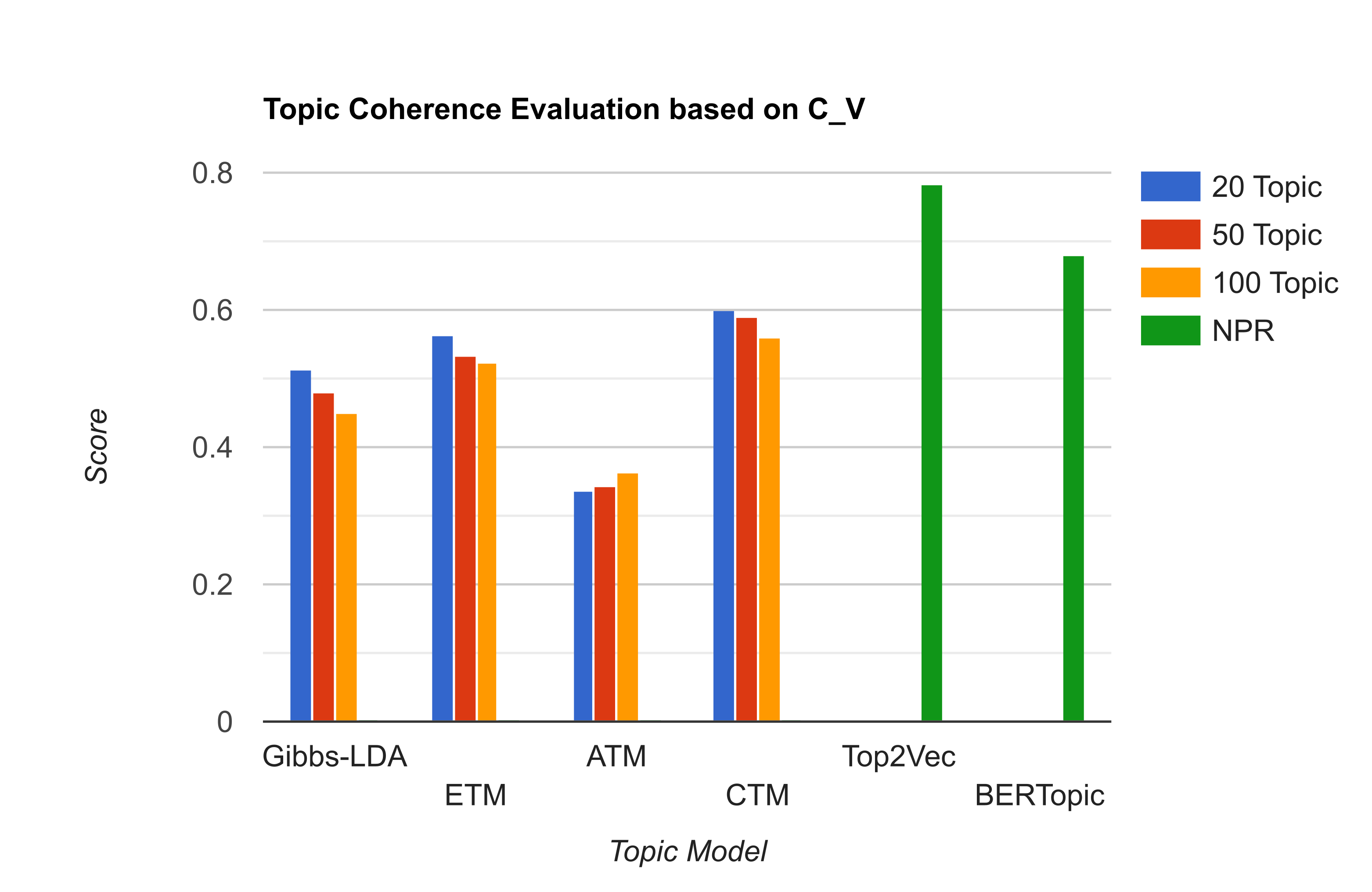}
\caption{20Newsgroup | C$_\text{V}$}
\label{fig:comp-a}
\end{subfigure}
\begin{subfigure}[b]{0.45\textwidth}
\includegraphics[width=\textwidth]{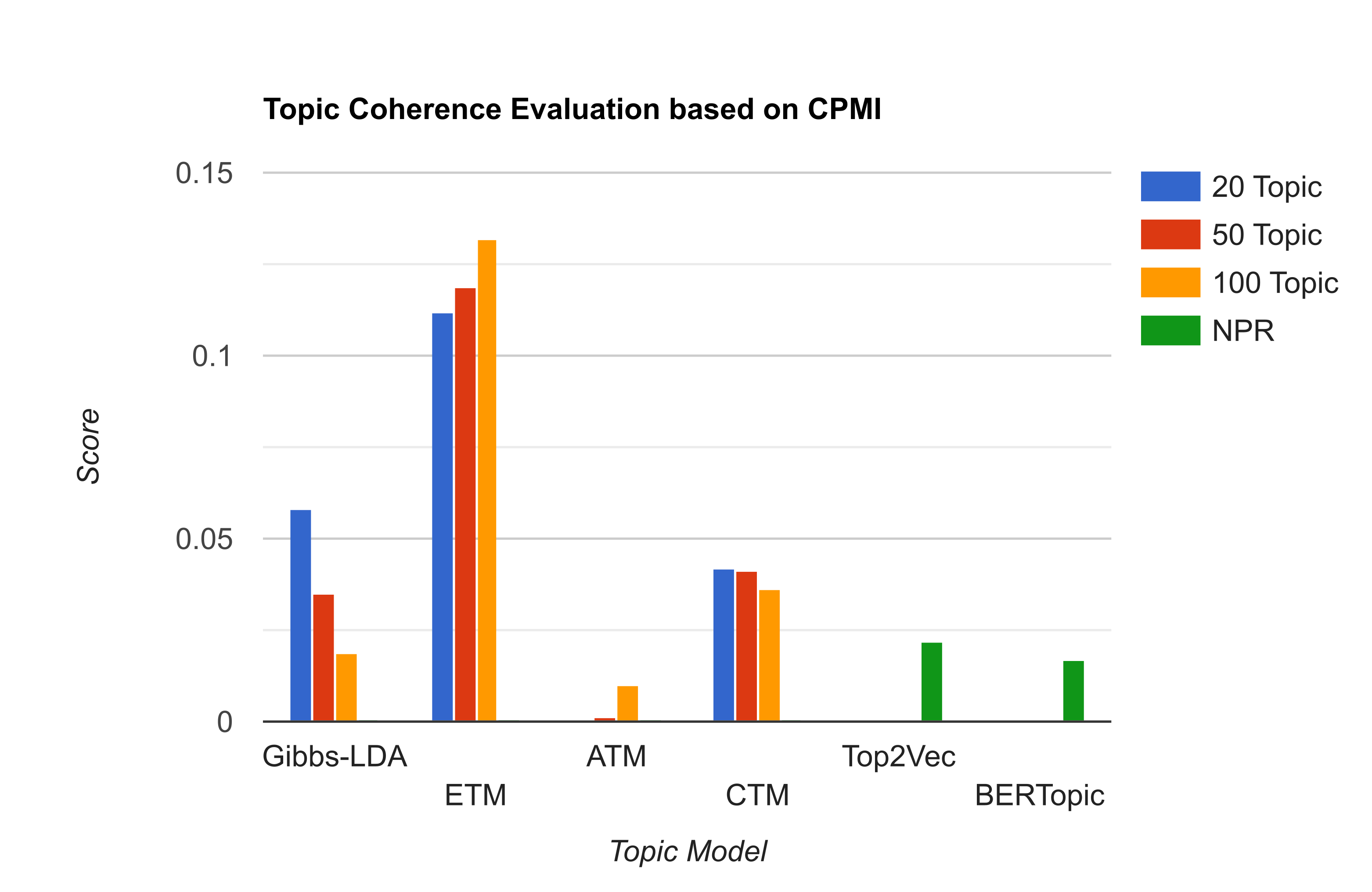}
\caption{20Newsgroup | CPMI}
\label{fig:comp-b}
\end{subfigure}
\begin{subfigure}[b]{0.45\textwidth}
\includegraphics[width=\textwidth]{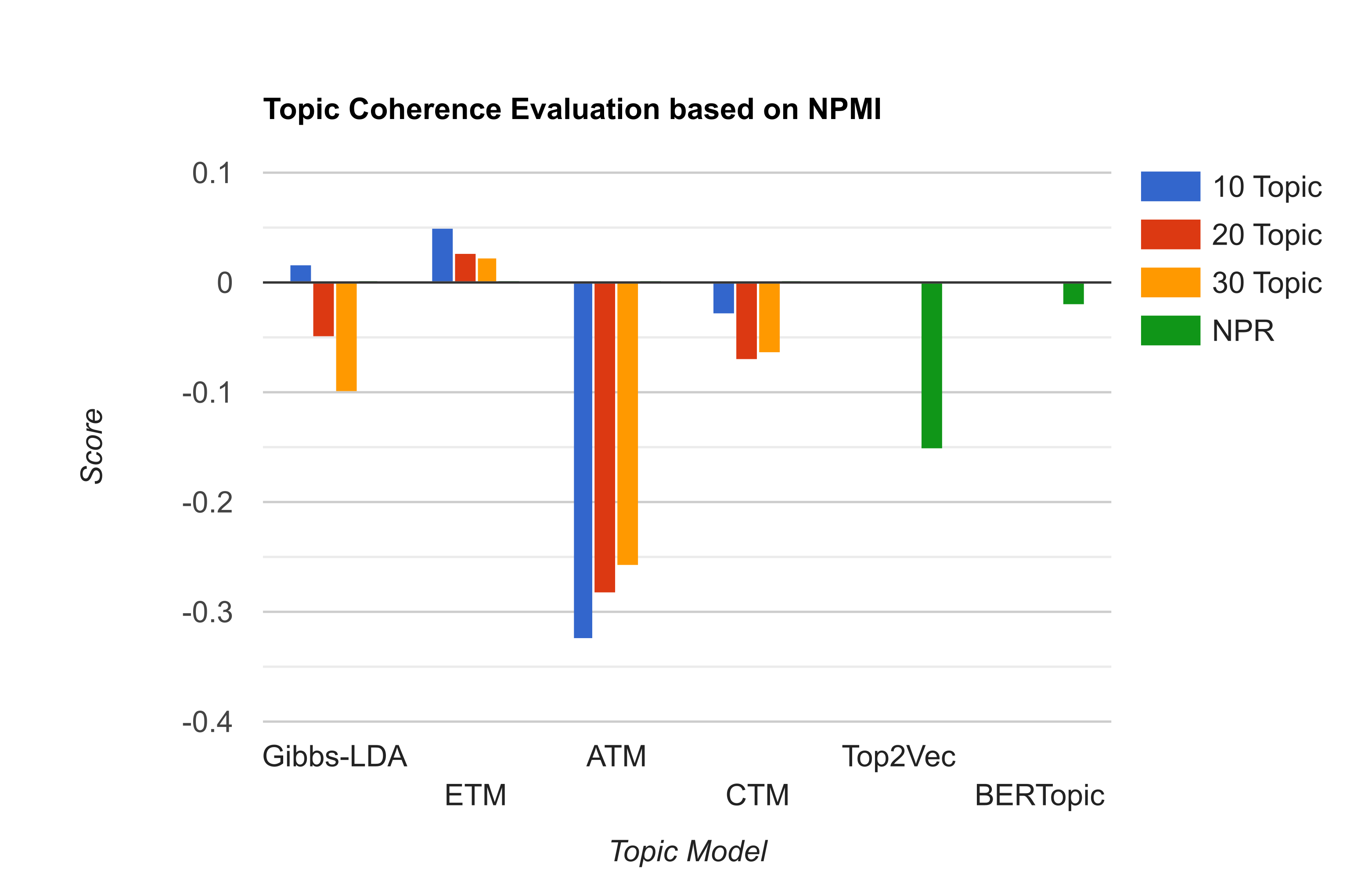}
\caption{Elon Musk's Tweets | NPMI}
\label{fig:comp-c}
\end{subfigure}
\begin{subfigure}[b]{0.45\textwidth}
\includegraphics[width=\textwidth]{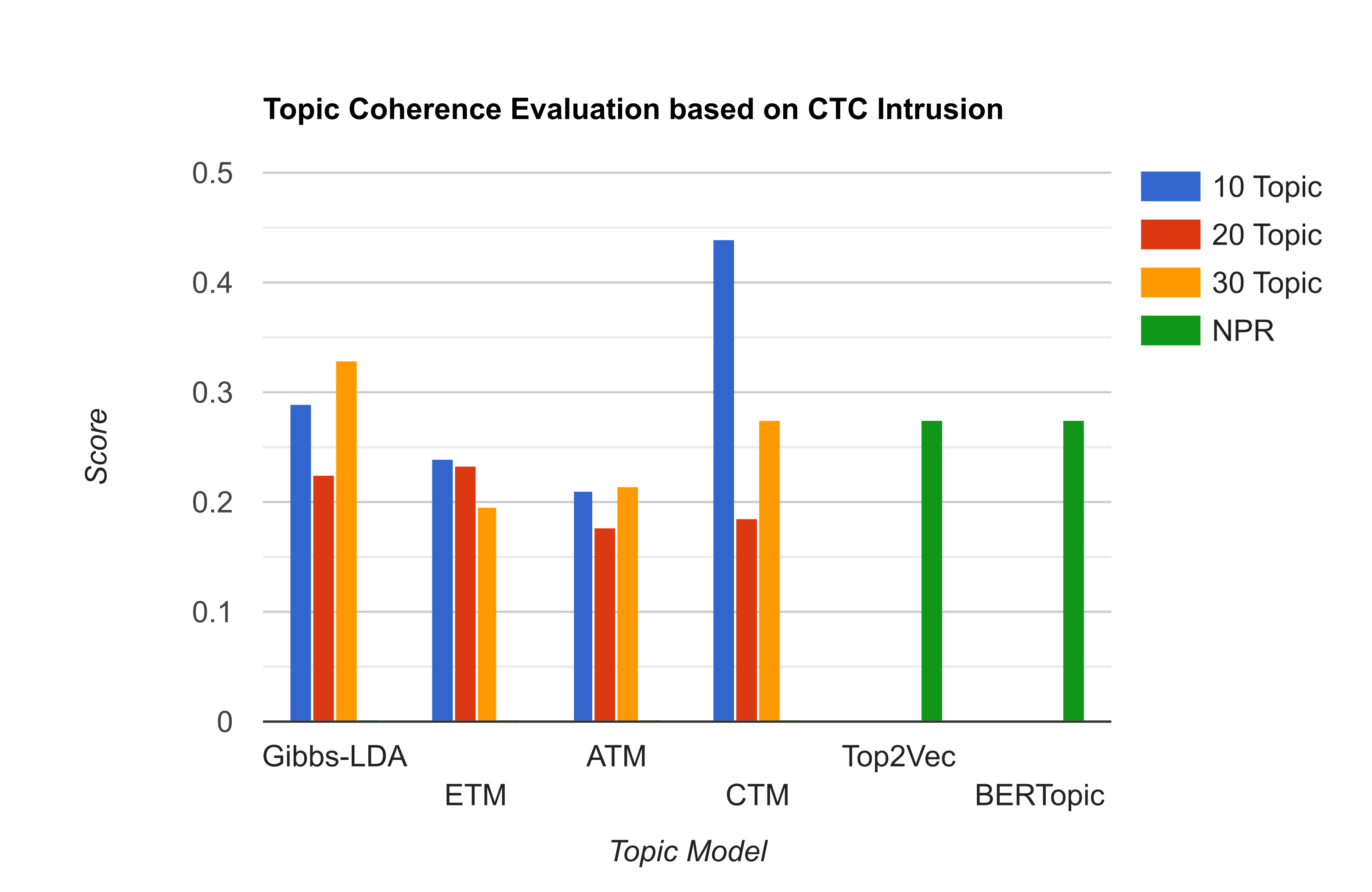}
\caption{Elon Musk's Tweets | Intrusion}
\label{fig:comp-d}
\end{subfigure}
\caption{Comparison Between Topic Models based on Topic Coherence Evaluation}
\label{fig:comp}
\end{figure}

\begin{table}[htb]
\scriptsize
\centering
\caption{Top-2 and bottom-2 topics of ETM$^{(100)}$ and Top2Vec scored by C$_V$ and CTC$_{\text{CPMI}}$ on 20 Newsgroup}
\label{tab:topic_comparison_20}
 \begin{tabular}{ |l|l|l||ll| }
\hline
Topic Model & Ranked By & Topics & C$_V$ & CPMI \\ 
\hline
\multirow{12}{*}{ETM$^{(100)}$ \citeyearpar{dieng2020topic}} 
& \multirow{3}{*}{Highest C$_\text{V}$}  
& god, christian, people, believe, jesus & 0.740 & 0.017 \\
& &drive, card, scsi, disk, mb, & 0.739 & 0.037  \\
\cline{2-5}
& \multirow{3}{*}{Lowest C$_\text{V}$}  
& book, number, problem, read, call & 0.369 & 0.018  \\
& &line, use, power, bit, high & 0.458 & 0.018  \\
\cline{2-5}
& \multirow{3}{*}{Highest CPMI}  
& year, time, day, one, ago, week & 0.559 & 0.709  \\
& & game, year, team, player, play & 0.706 & 0.242\\
\cline{2-5}
& \multirow{3}{*}{Lowest CPMI}  
&  new, number, also, well, call, order, used & 0.340 & -0.007  \\
& & people, right, drug, state, world, country  & 0.529 & -0.002 \\
\hline
\hline
\multirow{12}{*}{Top2Vec \citeyearpar{angelov2020top2vec}} 
& \multirow{3}{*}{Highest C$_\text{V}$}  
& dsl, geb, cadre, shameful, jxp & 0.995 & 0.009 \\
& & tor, nyi, det, chi, bos  & 0.989 & 0.012  \\
\cline{2-5}
& \multirow{3}{*}{Lowest C$_\text{V}$}  
& hacker, computer, privacy, uci, ethic & 0.255 & -0.0001  \\
& & battery, acid, charged, storage, floor  & 0.344 & 0.006  \\
\cline{2-5}
& \multirow{3}{*}{Highest CPMI}  
& mailing, list, mail, address, send & 0.792	& 0.154  \\
& & icon, window, manager, file, application  & 0.770 & 0.076 \\
\cline{2-5}
& \multirow{3}{*}{Lowest CPMI}  
& lc, lciii, fpu, slot, nubus, iisi & 0.853 & -0.004 \\
& & ci, ic, incoming, gif, edu & 0.644 & -0.002 \\
\hline
\end{tabular}
\end{table}
To verify the consistency of some representative scores in \Cref{tab:20Newsgroup}, we examine the topics for 2O Newsgroup generated by Top2Vec, which have high and low scores for baseline metrics, and ETM, which have high and low scores for CTC metrics. \Cref{tab:topic_comparison_20} compares the top-2 and bottom-2 topics ranked by C$_\text{V}$ and CTC$_{\text{CPMI}}$. The motivation behind choosing these metrics is from our correlation analysis in \Cref{fig:cor-a}, which in CTC$_{\text{CPMI}}$ and C$_\text{V}$ has the least correlation among CTC and baseline metrics. First, we notice that the top-2 topics returned by C$_\text{V}$ for Top2Vec are not readily interpretable but are statistically meaningful: \textit{dsl, geb, cadre, shameful, jxp} are fragments of an email signature that occurs 82 times, while \textit{tor, nyi, det, chi, bos} are abbreviations for hockey teams. This is not surprising, since Top2Vec produces what we call ``trash topics'', which is a common problem for clustering-based topic models that cannot handle so-called ``trash clusters''~\citep{giannotti2002clustering}. While CTC$_{\text{CPMI}}$ returns a more coherent ranking for Top2Vec (the top 2 topics appear coherent, while the bottom topics are incoherent for human evaluation). This proves our assumption that traditional topic coherence metrics such as C$_V$ might fail to evaluate neural topic models and, in this case, even give the highest scores to trash topics. This happens because they only consider the syntactic co-occurrence of words in a window of text and cannot observe the underlying relationship between topic words.  CTC$_{\text{CPMI}}$, on the other hand, can detect these trash topics and score them more accurately because it is contextual and accompanied by LLMs that have rich information about linguistic dependencies between topic words. CTC$_{\text{CPMI}}$ then also might be a good measure to filter out these topics. The second observation in \Cref{tab:topic_comparison_20} is that all eight topics returned for ETM are coherent. This is because ETM, which is a semantically-enabled probabilistic topic model, produces decent topics that are overall highly ranked by CTC$_{\text{CPMI}}$, as shown in \Cref{fig:comp-b}. 
%
%

\begin{table}[htbp]
\scriptsize
\centering
\caption{Top-2 and bottom-2 topics of ETM$^{(30)}$ and CTM$^{(30)}$ scored by NPMI and CTC$_{\text{Rating}}$ and CTC$_{\text{Intrusion}}$ on Elon Musk's Tweets}
\label{tab:topic_comparison_EM}
 \begin{tabular}{ |l|l|l||lll| }
\hline
Topic Model & Ranked By & Topics & NPMI & Rating & Intrusion \\ 
\hline
\multirow{12}{*}{CTM$^{(30)}$ \citeyearpar{bianchi-etal-2021-pre}} 
& \multirow{3}{*}{Highest NPMI}  
& erdayastronaut, engine, booster, starship, amp & 0.122 & 3 & 0.1  \\
& & year, week, next, month, wholemarsblog& 0.057 & 2 & 0.1 \\
\cline{2-6}
& \multirow{3}{*}{Lowest NPMI}  
& transport, backup, ensure, installed, transaction & -0.480 & 2 & 0.1 \\
& & achieving, transition, late, transport, precision & -0.459 & 1 & 0.1  \\
\cline{2-6}
& \multirow{3}{*}{Highest Rating}  
& tesla, rt, model, car, supercharger & -0.152 & 3 & 0.5  \\
& & spacex, dragon, launch, falcon, nasa & -0.283 & 3 & 0.4 \\
\cline{2-6}
& \multirow{3}{*}{Lowest Rating}  
& ppathole, soon, justpaulinelol, yes, sure  & -0.330 & 1 & 0.5 \\
& & achieving, transition, late, transport, precision & -0.459 & 1 & 0.1   \\
\hline
\hline
\multirow{12}{*}{ETM$^{(30)}$ \citeyearpar{dieng2020topic}} 
& \multirow{3}{*}{Highest NPMI}  
& amp, time, people, like, would, many & 0.001 & 2 & 0.7  \\
& & engine, booster, starship, heavy, raptor & -0.023 & 2 & 0.1   \\
\cline{2-6}
& \multirow{3}{*}{Lowest NPMI}  
& amp, rt, tesla, im, yes & -0.283 & 1 & 0.1   \\
& & amp, tesla, year, twitter, work & -0.228 & 1 & 0.1   \\
\cline{2-6}
& \multirow{3}{*}{Highest Rating}  
& amp, twitter, like, tesla, dont & -0.186 & 2 & 0.8   \\
& & amp, time, people, like, would & 0.001 & 2 & 0.7 \\
\cline{2-6}
& \multirow{3}{*}{Lowest Rating}  
& amp, tesla, year, twitter, work & -0.228 & 1 & 0.1   \\
& & amp, tesla, one, like, time & -0.204 & 1 & 0.1   \\
\hline
\end{tabular}
\end{table}
In the same way, we check the consistency of some representative scores in \Cref{tab:EM_Tweets} by checking the interpretability of topics for Elon Musk's tweets generated by ETM, which has high baseline scores, and by CTM, which has high CTC scores. As shown in \Cref{tab:topic_comparison_EM}, we compare the top 2 and bottom 2 topics ranked by NPMI and $\text{CTC}_\text{Rating}$. As shown in \Cref{fig:cor-b}, these metrics are among those with the lowest correlation between CTC and baseline metrics. 
A notable finding for CTM topics is that topics ranked highest by the $\text{CTC}_\text{Rating}$ metric tend to be more interpretable compared to those ranked highest by NPMI, and similarly, topics ranked lowest by the $\text{CTC}_\text{Rating}$ metric tend to be less interpretable compared to those ranked lowest by NPMI. 
The above observation also holds true for ETM, as the $\text{CTC}_\text{Rating}$ metric is not affected by the scarcity of short text records. This is because $\text{CTC}_\text{Rating}$ is complemented by a chatbot that mitigates the impact of limited data availability. It is also interesting to note that the topics generated by CTM are overall more interpretable and coherent than those generated by ETM. This demonstrates the validity of $\text{CTC}_\text{Rating}$ and $\text{CTC}_\text{Intrusion}$ over baseline metrics, as we observed in \Cref{tab:EM_Tweets}. It also reveals the superiority of CTM over ETM, as shown in \Cref{fig:comp-d}, in short text datasets as a result of a contextualized element in its architecture. 

\section{Conclusion}
This paper introduces a new family of topic coherence metrics called Contextualized Topic Coherence Metrics (CTC) that benefits from the recent development of Large Language Models (LLM). CTC includes two approaches that are motivated to offer flexibility and accuracy in evaluating neural topic models under different circumstances. Our results show automated CTC outperforms the baseline metrics on large-scale datasets while semi-automated CTC outperforms the baseline metrics on short-text datasets. After a comprehensive comparison between recent neural topic models and dominant classical topic models, the results indicate that some neural topic models, which optimize traditional topic coherence metrics, often manipulate traditional metrics and receive high scores while they produce meaningless and uninterpretable topics. We show with our experiments that CTC is not susceptible to being deceived by these meaningless topics since it has a strong ability to understand language and evaluate topics within their contextual framework. 

\newpage
\begin{ack}
    We gratefully acknowledge the Sorbonne Center for Artificial Intelligence (SCAI) for partially funding this research through a doctoral fellowship grant. Their support has been instrumental in enabling the successful execution of this study. 
\end{ack}
\bibliographystyle{abbrvnat}
\bibliography{ref}

\begin{thebibliography}{47}
\providecommand{\natexlab}[1]{#1}
\providecommand{\url}[1]{\texttt{#1}}
\expandafter\ifx\csname urlstyle\endcsname\relax
  \providecommand{\doi}[1]{doi: #1}\else
  \providecommand{\doi}{doi: \begingroup \urlstyle{rm}\Url}\fi

\bibitem[Abdelrazek et~al.(2022)Abdelrazek, Eid, Gawish, Medhat, and
  Hassan]{abdelrazek2022topic}
A.~Abdelrazek, Y.~Eid, E.~Gawish, W.~Medhat, and A.~Hassan.
\newblock Topic modeling algorithms and applications: A survey.
\newblock \emph{Information Systems}, page 102131, 2022.

\bibitem[Aletras and Stevenson(2013)]{aletras2013evaluating}
N.~Aletras and M.~Stevenson.
\newblock Evaluating topic coherence using distributional semantics.
\newblock In \emph{Proceedings of the 10th international conference on
  computational semantics (IWCS 2013)--Long Papers}, pages 13--22, 2013.

\bibitem[Angelov(2020)]{angelov2020top2vec}
D.~Angelov.
\newblock Top2vec: Distributed representations of topics.
\newblock \emph{arXiv preprint arXiv:2008.09470}, 2020.

\bibitem[Bianchi et~al.(2021)Bianchi, Terragni, and
  Hovy]{bianchi-etal-2021-pre}
F.~Bianchi, S.~Terragni, and D.~Hovy.
\newblock Pre-training is a hot topic: Contextualized document embeddings
  improve topic coherence.
\newblock In \emph{Proceedings of the 59th Annual Meeting of the Association
  for Computational Linguistics and the 11th International Joint Conference on
  Natural Language Processing (Volume 2: Short Papers)}, pages 759--766,
  Online, Aug. 2021. Association for Computational Linguistics.
\newblock \doi{10.18653/v1/2021.acl-short.96}.
\newblock URL \url{https://aclanthology.org/2021.acl-short.96}.

\bibitem[Bisong and Bisong(2019)]{bisong2019google}
E.~Bisong and E.~Bisong.
\newblock Google colaboratory.
\newblock \emph{Building machine learning and deep learning models on google
  cloud platform: a comprehensive guide for beginners}, pages 59--64, 2019.

\bibitem[Blei(2012)]{blei2012probabilistic}
D.~M. Blei.
\newblock Probabilistic topic models.
\newblock \emph{Communications of the ACM}, 55\penalty0 (4):\penalty0 77--84,
  2012.

\bibitem[Blei et~al.(2003)Blei, Ng, and Jordan]{blei2003latent}
D.~M. Blei, A.~Y. Ng, and M.~I. Jordan.
\newblock Latent dirichlet allocation.
\newblock \emph{Journal of machine Learning research}, 3\penalty0
  (Jan):\penalty0 993--1022, 2003.

\bibitem[Bouma(2009)]{bouma2009normalized}
G.~Bouma.
\newblock Normalized (pointwise) mutual information in collocation extraction.
\newblock \emph{Proceedings of GSCL}, 30:\penalty0 31--40, 2009.

\bibitem[Campagnolo et~al.(2022)Campagnolo, Duarte, and
  Dal~Bianco]{campagnolo2022topic}
J.~M. Campagnolo, D.~Duarte, and G.~Dal~Bianco.
\newblock Topic coherence metrics: How sensitive are they?
\newblock \emph{Journal of Information and Data Management}, 13\penalty0 (4),
  2022.

\bibitem[Chang et~al.(2009)Chang, Gerrish, Wang, Boyd-Graber, and
  Blei]{chang2009reading}
J.~Chang, S.~Gerrish, C.~Wang, J.~Boyd-Graber, and D.~Blei.
\newblock Reading tea leaves: How humans interpret topic models.
\newblock \emph{Advances in neural information processing systems}, 22, 2009.

\bibitem[Churchill and Singh(2022)]{churchill2022evolution}
R.~Churchill and L.~Singh.
\newblock The evolution of topic modeling.
\newblock \emph{ACM Computing Surveys}, 54\penalty0 (10s):\penalty0 1--35,
  2022.

\bibitem[Cover(1999)]{cover1999elements}
T.~M. Cover.
\newblock \emph{Elements of information theory}.
\newblock John Wiley \& Sons, 1999.

\bibitem[Dieng et~al.(2020)Dieng, Ruiz, and Blei]{dieng2020topic}
A.~B. Dieng, F.~J. Ruiz, and D.~M. Blei.
\newblock Topic modeling in embedding spaces.
\newblock \emph{Transactions of the Association for Computational Linguistics},
  8:\penalty0 439--453, 2020.

\bibitem[Doogan and Buntine(2021)]{doogan-buntine-2021-topic}
C.~Doogan and W.~Buntine.
\newblock Topic model or topic twaddle? re-evaluating semantic interpretability
  measures.
\newblock In \emph{Proceedings of the 2021 Conference of the North American
  Chapter of the Association for Computational Linguistics: Human Language
  Technologies}, pages 3824--3848, Online, June 2021. Association for
  Computational Linguistics.
\newblock \doi{10.18653/v1/2021.naacl-main.300}.
\newblock URL \url{https://aclanthology.org/2021.naacl-main.300}.

\bibitem[Giannotti et~al.(2002)Giannotti, Gozzi, and
  Manco]{giannotti2002clustering}
F.~Giannotti, C.~Gozzi, and G.~Manco.
\newblock Clustering transactional data.
\newblock In \emph{Principles of Data Mining and Knowledge Discovery: 6th
  European Conference, PKDD 2002 Helsinki, Finland, August 19--23, 2002
  Proceedings 6}, pages 175--187. Springer, 2002.

\bibitem[Griffiths and Steyvers(2004)]{griffiths2004finding}
T.~L. Griffiths and M.~Steyvers.
\newblock Finding scientific topics.
\newblock \emph{Proceedings of the National academy of Sciences}, 101\penalty0
  (suppl\_1):\penalty0 5228--5235, 2004.

\bibitem[Grootendorst(2022)]{grootendorst2022bertopic}
M.~Grootendorst.
\newblock Bertopic: Neural topic modeling with a class-based tf-idf procedure.
\newblock \emph{arXiv preprint arXiv:2203.05794}, 2022.

\bibitem[Harrando et~al.(2021)Harrando, Lisena, and Troncy]{harrando2021apples}
I.~Harrando, P.~Lisena, and R.~Troncy.
\newblock Apples to apples: A systematic evaluation of topic models.
\newblock In \emph{Proceedings of the International Conference on Recent
  Advances in Natural Language Processing (RANLP 2021)}, pages 483--493, 2021.

\bibitem[Hoover et~al.(2021)Hoover, Du, Sordoni, and
  O{'}Donnell]{hoover2021linguistic}
J.~L. Hoover, W.~Du, A.~Sordoni, and T.~J. O{'}Donnell.
\newblock Linguistic dependencies and statistical dependence.
\newblock In \emph{Proceedings of the 2021 Conference on Empirical Methods in
  Natural Language Processing}, pages 2941--2963, Online and Punta Cana,
  Dominican Republic, Nov. 2021. Association for Computational Linguistics.
\newblock \doi{10.18653/v1/2021.emnlp-main.234}.
\newblock URL \url{https://aclanthology.org/2021.emnlp-main.234}.

\bibitem[Hoyle et~al.(2021)Hoyle, Goel, Hian-Cheong, Peskov, Boyd-Graber, and
  Resnik]{hoyle2021automated}
A.~Hoyle, P.~Goel, A.~Hian-Cheong, D.~Peskov, J.~Boyd-Graber, and P.~Resnik.
\newblock Is automated topic model evaluation broken? the incoherence of
  coherence.
\newblock \emph{Advances in Neural Information Processing Systems},
  34:\penalty0 2018--2033, 2021.

\bibitem[Hoyle et~al.(2022)Hoyle, Goel, Sarkar, and Resnik]{hoyle2022neural}
A.~Hoyle, P.~Goel, R.~Sarkar, and P.~Resnik.
\newblock Are neural topic models broken?
\newblock \emph{arXiv preprint arXiv:2210.16162}, 2022.

\bibitem[Koren{\v{c}}i{\'c} et~al.(2018)Koren{\v{c}}i{\'c}, Ristov, and
  {\v{S}}najder]{korenvcic2018document}
D.~Koren{\v{c}}i{\'c}, S.~Ristov, and J.~{\v{S}}najder.
\newblock Document-based topic coherence measures for news media text.
\newblock \emph{Expert systems with Applications}, 114:\penalty0 357--373,
  2018.

\bibitem[Lang(1995)]{Lang95}
K.~Lang.
\newblock Newsweeder: Learning to filter netnews.
\newblock In \emph{Proceedings of the Twelfth International Conference on
  Machine Learning}, pages 331--339, 1995.

\bibitem[Lau et~al.(2014)Lau, Newman, and Baldwin]{lau2014machine}
J.~H. Lau, D.~Newman, and T.~Baldwin.
\newblock Machine reading tea leaves: Automatically evaluating topic coherence
  and topic model quality.
\newblock In \emph{Proceedings of the 14th Conference of the European Chapter
  of the Association for Computational Linguistics}, pages 530--539, 2014.

\bibitem[Le and Mikolov(2014)]{le2014distributed}
Q.~Le and T.~Mikolov.
\newblock Distributed representations of sentences and documents.
\newblock In \emph{International conference on machine learning}, pages
  1188--1196. PMLR, 2014.

\bibitem[Lund et~al.(2019)Lund, Armstrong, Fearn, Cowley, Byun, Boyd-Graber,
  and Seppi]{lund2019automatic}
J.~Lund, P.~Armstrong, W.~Fearn, S.~Cowley, C.~Byun, J.~Boyd-Graber, and
  K.~Seppi.
\newblock Automatic evaluation of local topic quality.
\newblock \emph{arXiv preprint arXiv:1905.13126}, 2019.

\bibitem[McCallum(2002)]{mccallum2002mallet}
A.~K. McCallum.
\newblock Mallet: A machine learning for languagetoolkit.
\newblock \emph{http://mallet. cs. umass. edu}, 2002.

\bibitem[Mikolov et~al.(2013{\natexlab{a}})Mikolov, Chen, Corrado, and
  Dean]{mikolov2013efficient}
T.~Mikolov, K.~Chen, G.~Corrado, and J.~Dean.
\newblock Efficient estimation of word representations in vector space.
\newblock \emph{arXiv preprint arXiv:1301.3781}, 2013{\natexlab{a}}.

\bibitem[Mikolov et~al.(2013{\natexlab{b}})Mikolov, Sutskever, Chen, Corrado,
  and Dean]{mikolov2013distributed}
T.~Mikolov, I.~Sutskever, K.~Chen, G.~S. Corrado, and J.~Dean.
\newblock Distributed representations of words and phrases and their
  compositionality.
\newblock \emph{Advances in neural information processing systems}, 26,
  2013{\natexlab{b}}.

\bibitem[Mimno et~al.(2011)Mimno, Wallach, Talley, Leenders, and
  McCallum]{mimno2011optimizing}
D.~Mimno, H.~Wallach, E.~Talley, M.~Leenders, and A.~McCallum.
\newblock Optimizing semantic coherence in topic models.
\newblock In \emph{Proceedings of the 2011 conference on empirical methods in
  natural language processing}, pages 262--272, 2011.

\bibitem[Newman et~al.(2009)Newman, Karimi, and Cavedon]{NewmanDavid2009Eeot}
D.~Newman, S.~Karimi, and L.~Cavedon.
\newblock External evaluation of topic models.
\newblock In \emph{Proceedings of the 14th Australasian Document Computing
  Symposium}, pages 1--8. University of Sydney, 2009.

\bibitem[Newman et~al.(2010{\natexlab{a}})Newman, Lau, Grieser, and
  Baldwin]{newman2010automatic}
D.~Newman, J.~H. Lau, K.~Grieser, and T.~Baldwin.
\newblock Automatic evaluation of topic coherence.
\newblock In \emph{Human language technologies: The 2010 annual conference of
  the North American chapter of the association for computational linguistics},
  pages 100--108, 2010{\natexlab{a}}.

\bibitem[Newman et~al.(2010{\natexlab{b}})Newman, Noh, Talley, Karimi, and
  Baldwin]{10.1145/1816123.1816156}
D.~Newman, Y.~Noh, E.~Talley, S.~Karimi, and T.~Baldwin.
\newblock Evaluating topic models for digital libraries.
\newblock In \emph{Proceedings of the 10th Annual Joint Conference on Digital
  Libraries}, JCDL '10, page 215–224, New York, NY, USA, 2010{\natexlab{b}}.
  Association for Computing Machinery.
\newblock ISBN 9781450300858.
\newblock \doi{10.1145/1816123.1816156}.
\newblock URL \url{https://doi.org/10.1145/1816123.1816156}.

\bibitem[Nikolenko(2016)]{10.1145/2911451.2914720}
S.~I. Nikolenko.
\newblock Topic quality metrics based on distributed word representations.
\newblock In \emph{Proceedings of the 39th International ACM SIGIR Conference
  on Research and Development in Information Retrieval}, SIGIR '16, page
  1029–1032, New York, NY, USA, 2016. Association for Computing Machinery.
\newblock ISBN 9781450340694.
\newblock \doi{10.1145/2911451.2914720}.
\newblock URL \url{https://doi.org/10.1145/2911451.2914720}.

\bibitem[OpenAI(2022)]{openai2022chatgpt}
OpenAI.
\newblock Chatgpt: Engaging and dynamic conversations.
\newblock \url{https://openai.com/blog/chatgpt}, 2022.

\bibitem[Rahimi et~al.(2023)Rahimi, Naacke, Constantin, and
  Amann]{rahimi2023antm}
H.~Rahimi, H.~Naacke, C.~Constantin, and B.~Amann.
\newblock Antm: An aligned neural topic model for exploring evolving topics.
\newblock \emph{arXiv preprint arXiv:2302.01501}, 2023.

\bibitem[Ramrakhiyani et~al.(2017)Ramrakhiyani, Pawar, Hingmire, and
  Palshikar]{ramrakhiyani2017measuring}
N.~Ramrakhiyani, S.~Pawar, S.~Hingmire, and G.~Palshikar.
\newblock Measuring topic coherence through optimal word buckets.
\newblock In \emph{Proceedings of the 15th Conference of the European Chapter
  of the Association for Computational Linguistics: Volume 2, Short Papers},
  pages 437--442, 2017.

\bibitem[Raza(2023)]{raza2023elon}
Y.~Raza.
\newblock Elon musk tweets dataset (17k): Dataset of elon musk tweets till now
  (17k).
\newblock
  \url{https://www.kaggle.com/datasets/yasirabdaali/elon-musk-tweets-dataset-17k},
  2023.

\bibitem[R{\"o}der et~al.(2015)R{\"o}der, Both, and
  Hinneburg]{roder2015exploring}
M.~R{\"o}der, A.~Both, and A.~Hinneburg.
\newblock Exploring the space of topic coherence measures.
\newblock In \emph{Proceedings of the eighth ACM international conference on
  Web search and data mining}, pages 399--408, 2015.

\bibitem[Schnabel et~al.(2015)Schnabel, Labutov, Mimno, and
  Joachims]{schnabel2015evaluation}
T.~Schnabel, I.~Labutov, D.~Mimno, and T.~Joachims.
\newblock Evaluation methods for unsupervised word embeddings.
\newblock In \emph{Proceedings of the 2015 conference on empirical methods in
  natural language processing}, pages 298--307, 2015.

\bibitem[Sedgwick(2012)]{sedgwick2012pearson}
P.~Sedgwick.
\newblock Pearson’s correlation coefficient.
\newblock \emph{Bmj}, 345, 2012.

\bibitem[Srivastava and Sutton(2017)]{srivastava2017autoencoding}
A.~Srivastava and C.~Sutton.
\newblock Autoencoding variational inference for topic models.
\newblock \emph{arXiv preprint arXiv:1703.01488}, 2017.

\bibitem[Stevens et~al.(2012)Stevens, Kegelmeyer, Andrzejewski, and
  Buttler]{stevens2012exploring}
K.~Stevens, P.~Kegelmeyer, D.~Andrzejewski, and D.~Buttler.
\newblock Exploring topic coherence over many models and many topics.
\newblock In \emph{Proceedings of the 2012 joint conference on empirical
  methods in natural language processing and computational natural language
  learning}, pages 952--961, 2012.

\bibitem[Syed and Spruit(2017)]{syed2017full}
S.~Syed and M.~Spruit.
\newblock Full-text or abstract? examining topic coherence scores using latent
  dirichlet allocation.
\newblock In \emph{2017 IEEE International conference on data science and
  advanced analytics (DSAA)}, pages 165--174. IEEE, 2017.

\bibitem[Thompson and Mimno(2020)]{thompson2020topic}
L.~Thompson and D.~Mimno.
\newblock Topic modeling with contextualized word representation clusters,
  2020.

\bibitem[Wang et~al.(2019)Wang, Zhou, and He]{wang2019atm}
R.~Wang, D.~Zhou, and Y.~He.
\newblock Atm: Adversarial-neural topic model.
\newblock \emph{Information Processing \& Management}, 56\penalty0
  (6):\penalty0 102098, 2019.

\bibitem[Zhao et~al.(2021)Zhao, Phung, Huynh, Jin, Du, and
  Buntine]{zhao2021topic}
H.~Zhao, D.~Phung, V.~Huynh, Y.~Jin, L.~Du, and W.~Buntine.
\newblock Topic modelling meets deep neural networks: A survey.
\newblock \emph{arXiv preprint arXiv:2103.00498}, 2021.

\end{thebibliography}
\newpage

\appendix

\section{Normalized CPMI}
To improve comparability, we also propose a normalized version of CPMI that extend its generalizability and allows to mitigate potential biases that may arise due to specific dataset characteristics or idiosyncrasies. Additionally, it facilitates threshold determination and provides a consistent scale that allows researchers to set thresholds based on desired coherence levels, ensuring the metric is effectively utilized in practical applications.
\subsection{Definition}

 Given a set of $n$ topics $\text{TM}\mapsto\{t_1, t_2,\dots,t_n\}$ with $m$ words $t_i\mapsto\{w^i_1, w^i_2,\dots,w^i_m\}$ as an output of topic model $\text{TM}$ on the corpus of $e$ documents $D=\{d_1,d_2,\dots,d_e\}$, the CTC based on Normalized CPMI (NCPMI) called CTC$_{\text{NCPMI}}$ is defined as follows.

\begin{equation}
\text{CTC}_{\text{NCPMI}}=\frac{1}{e}\sum_{d=1}^{e}\frac{1}{n} \sum_{i=1}^{n}\frac{1}{m} \sum_{j=1}^{m}\text{NCPMI}(w^i_j,t^i\mid c^d)
\end{equation}
while:
\begin{equation}
\text{NCPMI}(w^i_j,t^i\mid c^d)= \frac{log\frac{P(w^i_j \mid c^d_{-w^i_j})}{P(w^i_j \mid c^d_{-t^i})}}{-log(P(w^i_j \mid c^d_{-w^i_j})\times P(t^i \mid c^d_{-t^i}))}  
\end{equation}
where $P$ is an estimate for the probability of words given context based on language model $\text{LM}$. The $c^d_{-w_i}$ is the document $d$ with word $w_i$ masked, and $c^d_{-t_j}$ is the document $d$ with words of topic $t^i$ masked.

\section{Reproducibility}
\subsection{Python Package}
CTC is implemented as a service for researchers and engineers who aim to evaluate and fine-tune their topic models. The source code of this python package is provided in \textit{./ctc} and a notebook named \textit{example.ipynb} is prepared to explain how to use this python package as follows.

\subsubsection{Automated CTC}

\begin{lstlisting}
from ctc.main import Auto_CTC
#initiating the metric
eval=Auto_CTC(segments_length=15, min_segment_length=5, segment_step=10,device="mps") 

# segmenting the documents
docs=documents 
eval.segmenting_documents(docs) 

# creating cpmi tree including all co-occurence values between all pairs of words 
eval.create_cpmi_tree() 
#eval.load_cpmi_tree() 

# topics=[["game","play"],["man","devil"]] for instance
eval.ctc_cpmi(topics) 
\end{lstlisting}

\subsubsection{Semi-automated CTC}

\begin{lstlisting}
from ctc.main import Semi_auto_CTC

openai_key="YOUR OPENAI KEY"

y=Semi_auto_CTC(openai_key,topics)

y.ctc_intrusion()

y.ctc_rating()
\end{lstlisting}

\subsection{Experiments}
The experiments of this paper including topic models, measuring scores, and complete results are provided in the \textit{./experiment} folder. To reproduce the results from scratch, it is required first to train topic models with the notebooks provided in \textit{./experiment/NTM}. Afterward, there are 4 notebooks in \textit{./experiment} to compute CTC$_{\text{CPMI}}$, CTC$_{\text{Intrusion}}$, CTC$_{\text{Rating}}$, and traditional topic coherence metrics and to save the results as \textit{.txt} files. The analysis is provided in the notebook named \textit{Analysis.ipynb}.

\end{document}